%% file: main.tex
\documentclass{IEEEtran}
\ifCLASSOPTIONcompsoc
  \usepackage[nocompress]{cite}
\else
  \usepackage{cite}
\fi
\ifCLASSINFOpdf
  \usepackage[pdftex]{graphicx}
\else
\fi
\usepackage{amsmath}

\usepackage{multirow}
\usepackage{array}

\begin{document}
%
\title{GAT-Steiner: Rectilinear Steiner Minimal Tree Prediction Using GNNs}

\author{\IEEEauthorblockN{Bugra Onal\IEEEauthorrefmark{2}, 
Eren Dogan\IEEEauthorrefmark{2}, 
Muhammad Hadir Khan,
Matthew R. Guthaus }\\
\IEEEauthorblockA{Computer Science and Engineering, University of California Santa Cruz, Santa Cruz, CA 95064\\ 
\{bonal, erdogan, mkhan33, mrg\}@ucsc.edu}}


%


\newcommand{\todo}[1]{{\bf TODO: {#1}}\\}

\input{variables}

\maketitle
\begingroup\renewcommand\thefootnote{\IEEEauthorrefmark{2}}
\footnotetext{Equal contribution}
\begingroup\renewcommand\thefootnote{\IEEEauthorrefmark{1}}

\begin{abstract}
The Rectilinear Steiner Minimum Tree (RSMT) problem is a fundamental problem in VLSI placement and routing and is known to be NP-hard. Traditional RSMT algorithms spend a significant amount of time on finding Steiner points to reduce the total wire length or use heuristics to approximate producing sub-optimal results. We show that Graph Neural Networks (GNNs) can be used to predict optimal Steiner points in RSMTs with high accuracy and can be parallelized on GPUs. In this paper, we propose GAT-Steiner, a graph attention network model that correctly predicts \modelIspdCorrectPredictions{} of the nets in the ISPD19 benchmark with an average increase in wire length of only \modelIspdAvgWLIncrease{} on suboptimal wire length nets. On randomly generated benchmarks, GAT-Steiner correctly predicts \modelRandomCorrectPredictions{}  with an average increase in wire length of only \modelRandomAvgWLIncrease{} on suboptimal wire length nets.
\end{abstract}


%
\IEEEpeerreviewmaketitle

\input{intro}

\input{background}
\input{implementation}

\input{results}

\input{conclusion}






%
\newpage
{\small
\bibliographystyle{ieeetr}
\bibliography{main}
}

\end{document}

%% file: variables.tex
\def \modelNumChannels {2}
\def \modelNumConvLayers {2}
\def \modelNumAttentionHeads {8}
\def \modelAttentionDropout {0.025}
\def \modelLayerDropout {0.0}
\def \trainingAcc {99.766\%} 

\def \fluteDefaultOpenroadAcc {99.949\%} 
\def \fluteDefaultOpenroadAvgWLIncrease {1.276\%} 
\def \fluteDefaultOpenroadMaxWLIncrease {5.773\%} 
\def \fluteDefaultOpenroadMinWLIncrease {0.037\%} 
\def \fluteBestOpenroadAcc {99.986\%} 
\def \fluteBestOpenroadAvgWLIncrease {0.758\%} 
\def \fluteBestOpenroadMaxWLIncrease {2.438\%} 
\def \fluteBestOpenroadMinWLIncrease {0.137\%} 
\def \saltOpenroadAcc {99.970\%} 
\def \saltOpenroadAvgWLIncrease {1.329\%} 
\def \saltOpenroadMaxWLIncrease {5.773\%} 
\def \saltOpenroadMinWLIncrease {0.218\%} 
\def \modelOpenroadAcc {99.889\%} 
\def \modelOpenroadAvgWLIncrease {1.050\%} 
\def \modelOpenroadMaxWLIncrease {15.863\%} 
\def \modelOpenroadMinWLIncrease {0.000\%} 
\def \fluteDefaultIspdAcc {95.592\%} 
\def \fluteDefaultIspdAvgWLIncrease {1.342\%} 
\def \fluteDefaultIspdMaxWLIncrease {23.276\%} 
\def \fluteDefaultIspdMinWLIncrease {0.000\%} 
\def \fluteDefaultIspdMispredictions {6.600\%}
\def \fluteBestIspdAcc {98.418\%} 
\def \fluteBestIspdAvgWLIncrease {0.470\%} 
\def \fluteBestIspdMaxWLIncrease {13.157\%} 
\def \fluteBestIspdMinWLIncrease {0.000\%} 
\def \fluteBestIspdMispredictions {2.640\%}
\def \saltIspdAcc {95.923\%} 
\def \saltIspdAvgWLIncrease {0.989\%} 
\def \saltIspdMaxWLIncrease {19.767\%} 
\def \saltIspdMinWLIncrease {0.000\%} 
\def \saltIspdMispredictions {5.770\%}
\def \modelIspdAcc {99.909\%} 
\def \modelIspdAvgWLIncrease {0.480\%} 
\def \modelIspdMaxWLIncrease {11.708\%} 
\def \modelIspdMinWLIncrease {0.001\%} 
\def \modelIspdMispredictions {0.154\%}
\def \modelIspdCorrectPredictions {99.846\%}

\def \fluteDefaultRandomAcc {52.071\%} 
\def \fluteDefaultRandomAvgWLIncrease {1.610\%} 
\def \fluteDefaultRandomMaxWLIncrease {11.585\%} 
\def \fluteDefaultRandomMinWLIncrease {0.000\%} 
\def \fluteDefaultRandomMispredictions {65.157\%}
\def \fluteBestRandomAcc {79.446\%} 
\def \fluteBestRandomAvgWLIncrease {0.516\%} 
\def \fluteBestRandomMaxWLIncrease {8.042\%} 
\def \fluteBestRandomMinWLIncrease {0.000\%} 
\def \fluteBestRandomMispredictions {29.784\%}
\def \saltRandomAcc {51.027\%} 
\def \saltRandomAvgWLIncrease {1.575\%} 
\def \saltRandomMaxWLIncrease {11.585\%} 
\def \saltRandomMinWLIncrease {0.000\%} 
\def \saltRandomMispredictions {63.884\%}
\def \modelRandomAcc {99.992\%} 
\def \modelRandomAvgWLIncrease {0.420\%} 
\def \modelRandomMaxWLIncrease {4.114\%} 
\def \modelRandomMinWLIncrease {0.003\%} 
\def \modelRandomMispredictions {0.058\%}
\def \modelRandomCorrectPredictions {99.942\%}

\def \modelSpeedupLargeBatch {9.363x}
\def \modelSpeedupSmallBatch {24.305x}

\def \modelRandomRanRefinement {2.254\%} 
\def \modelRandomRefinementAlreadyOptimal {2.107\%} 
\def \modelRandomAlreadyOptimalPercentage {93.478\%}
\def \modelIspdRanRefinement {1.471\%} 
\def \modelIspdRefinementAlreadyOptimal {1.274\%} 
\def \modelIspdAlreadyOptimalPercentage {85.608\%}

%% file: intro.tex
\section{Introduction}
\label{sec:intro}

To create a physical layout, placement and routing tools must connect together pins of nets with wires during the placement, routing and optimization phases.  Net routing, however, takes a significant amount of time since there are many such nets, so existing solutions typically make use of iterative heuristic approaches since the problem is known to be NP-hard. Long runtimes not only affect the time to market but also make it increasingly difficult to iterate in the design flow in the later stages.

In order to satisfy the power, performance, and area (PPA) requirements of a design, nets are typically connected with the shortest wire length possible. Modern placement and routing tools leverage Steiner points to achieve a better wire length through Steiner Minimum Trees (SMTs) or, a variant of this problem with orthogonal routing layers, Rectilinear Steiner Minimum Trees (RSMT). RSMT is an NP-hard problem, and finding these Steiner points accurately can be a time-consuming process for routing tools. Optimal RSMTs can be computed using Integer Linear Programming (ILP)~\cite{geosteiner}, but require significant run-time which limits their usage in placement and optimization steps.

There are a number of state-of-the-art RSMT heuristics such as FLUTE~\cite{flute} and SALT~\cite{salt}, but they give up wire-length optimality for run time. Another recent work~\cite{nnsteiner} utilizes a mixed neural network (NN) and dynamic programming (DP) approach using the PTAS algorithm~\cite{PTAS}, but does not use Graph Neural Networks (GNNs). Yet another deep reinforcement learning technique for non-rectilinear SMT problems has shown some promise but also relies on iterative Steiner point selection using GNNs~\cite{delaware}. In general, these algorithmic, iterative approaches limit batch computation and, at best, require multiple calls to a GPU for parallel computation.

We propose the first Graph Neural Network (GNN) model for the RSMT problem using a Graph Attention Network (GAT)~\cite{gat}. The idea is to directly predict Steiner points, while significantly reducing the runtime by computing these Steiner points in parallel rather than iteratively. In addition, implementations of GNNs can also predict the Steiner points for multiple Steiner trees at the same time using vector processing units offering speed-up at the design level.

In Section~\ref{sec:background}, we provide more context on the state-of-the-art techniques we use to achieve our high-accuracy model for RSMT prediction. In Section~\ref{sec:implementation}, we describe our model structure, how we implemented it and how we train it. In Section~\ref{sec:results}, we show how we tested the model, and analyze accuracy and parallelization we gained. Finally, Section~\ref{sec:conclusion} concludes the paper.

%% file: background.tex
\section{Background}
\label{sec:background}





GNNs are neural network (NN) techniques similar to Convolutional Neural Networks (CNNs); however, their input is a non-uniform graph instead of a regular matrix~\cite{gcn,gat,graphsage}. Technically, CNNs are a special, uniform graph case of GNNs. GNN models that interpret relations between nodes of the input graph can be trained to perform node prediction, edge prediction, and graph classification. 

GNNs can be trained for either transductive inference or inductive inference. Transductive inference refers to applications where the model is trained to solve a specific set of test inputs; whereas inductive learning refers to models trained to solve for a more general problem on unknown inputs. 

Similar to other NN methods, GNNs can be trained in both supervised and unsupervised fashions using back-propagation. Supervised learning is when the training data is labeled with known outputs so that patterns can be recognized between the inputs and outputs. With unsupervised training, the training data does not have labels and the model instead uses a loss heuristic, but these are often challenging to formulate for a given problem. 


A significant advantage that GNNs have over CNNs is that they can correctly predict outcomes for isomorphic graphs. If two graphs have a one-to-one mapping with different node ordering, they are isomorphic. In these two cases, CNNs will often predict different results for isomorphic graphs depending on training and node ordering, whereas GNNs predict consistently by aggregating features from topological neighbours. 

One issue with GNNs, however, is over-smoothing where node features tend to converge to similar values for deeply layered networks~\cite{oversmoothingsurvey}. This is mostly addressed by tuning the model to have appropriate levels, but has also been examined through adaptive mechanisms~\cite{jknet}. 

\begin{figure}[tb]
    \centering
    \includegraphics[width=\linewidth]{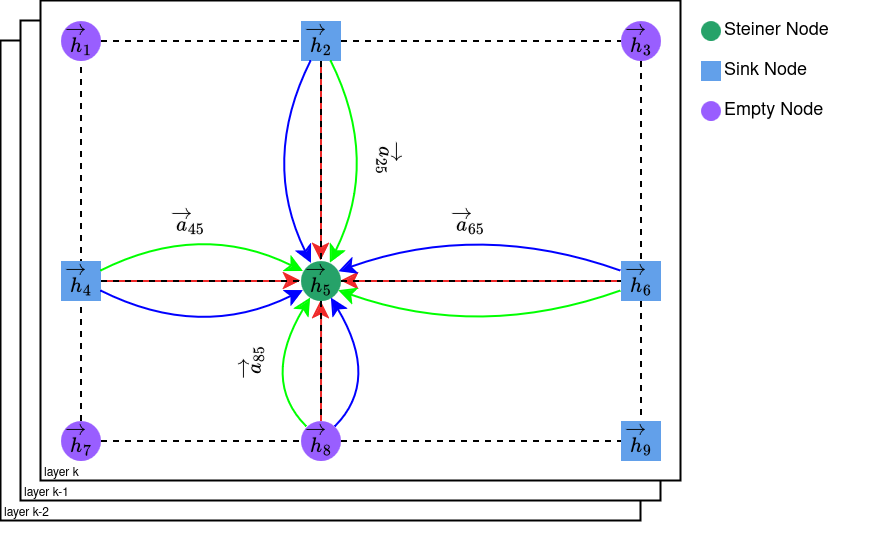}
    \caption{GAT message passing of nodes on a Hanan grid to update a single node's feature vector, $\overrightarrow{h}_5$. Three attention heads are shown (green, red, blue) and three layers are stacked on top of each other. Node features can be coordinates, node type, etc. and are aggregated with multiple-levels.}
    \label{fig:gnn-message-pass}
\end{figure}

The message passing feature is the core principle behind GNNs. It is an iterative process of updating the features of the nodes (i.e.,  $\overrightarrow{h}_{j}$ to $\overrightarrow{h}_{j+1}$) based on their neighbors' and their own features ($\overrightarrow{h}_i$ for $i \in \mathcal{N}_i$) using learnable weights.  Each iteration over all nodes is a GNN layer which allows features to be updated with information from neighbors that are one step further away. A message-passing layer can be written in matrix form as
\begin{equation}
    H_{k+1}=\sigma(A \times H_{k} \times W_k).
\end{equation}
where $H_{k}$ is matrix of feature vectors at the $k^{th}$ layer for all node features ($\overrightarrow{h}_i$); $A$ is the adjacency matrix of the graph; $W_k$ is the learnable component in the $k^{th}$ layer; and $\sigma(\cdot)$ is any non-linear activation function. This message-passing method is used as a way to discover feature embeddings for further use with downstream tasks such as node prediction.  An example of message passing on a Hanan grid graph for a single node is shown in Fig.~\ref{fig:gnn-message-pass}. 

Graph Attention Networks (GATs)~\cite{gat} are a specialized variant of GNNs, where the relations of neighbors can be learned with an attention mechanism~\cite{attention}.  GATs add an attention coefficient for each neighbor based on weighted feature similarity: 
\begin{equation}
    e_{ij} = \alpha(W{\overrightarrow{h}_i},W{\overrightarrow{h}_j}).
\end{equation}
These coefficients are made comparable at each layer and neighborhood by using the softmax of all neighbor attention coefficients, 
\begin{equation}
    \alpha_{ij}=softmax(e_{ij})
\end{equation}
as seen in Fig~\ref{fig:gnn-message-pass}.

GATs can also incorporate multiple attention heads to understand how portions of neighbor features may affect the output of that node differently~\cite{gat}.  Specifically, multiple weight matrices may be used and the weighted outputs combined using concatenation, averaging, or summation. For example, in rectilinear routing,  the left and up neighbors together may affect a node differently than the right or lower neighbors and may benefit from multiple attention heads. Fig.~\ref{fig:gnn-message-pass} shows three such attention heads in red, green, and blue entering the center node with different weights.

Dropout is a method used to prevent over-fitting the model for specific training data~\cite{dropout}. During training, a dropout layer randomly selects a set of input features with a predefined percentage and masks those out of the updates. Similarly, attention dropout in GNNs can select a predefined percentage of neighbors and mask those from a layer update during training. 

%% file: implementation.tex
\section{Implementation}
\label{sec:implementation} 
\begin{figure*}[tb] 
\begin{center}
\vspace{-0.3cm}
\includegraphics[width = \textwidth]{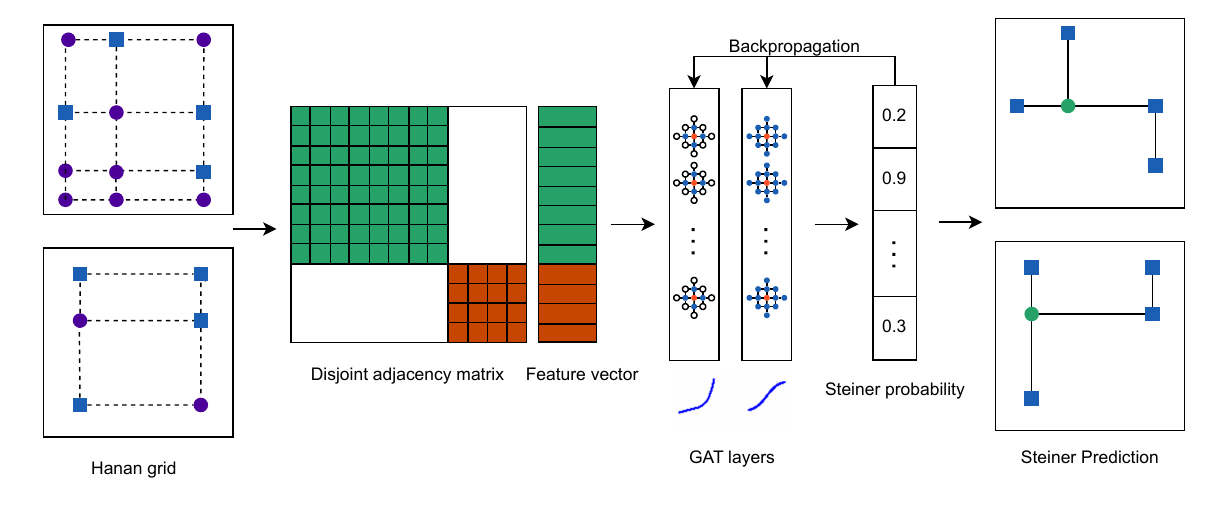}
 \caption {The model flow, starting with Hanan grids of nets; adjacency matrices and feature vectors are constructed. Then, feature embeddings are found using 2 GAT layers using multiple attention heads, with ELU and sigmoid activation functions respectively. Steiner point probability for each node is found and points above the threshold are selected as Steiner points. Finally, nets are routed by finding the MST of pins and Steiner nodes.
 \label{fig:model_structure}}
 \vspace{-0.3cm}
 \end{center}
\end{figure*}

The GAT-Steiner model is shown in Fig.~\ref{fig:model_structure} and is made up of a number of GAT convolutional layers~\cite{gat} configured to do node prediction. In addition, we implemented layer and attention dropout mechanisms. We adopt a supervised training model using optimal RSMTs generated by GeoSteiner but also use L2 regularization loss functions for kernel, bias and attention regularization of each layer. 

The output of the GAT model is the probability of each node being a Steiner node. We apply a threshold of 0.5 to select all the Steiner points from a single inference. Using the Steiner points along with the net pins, we route the net using Kruskal's MST algorithm. In this section, we discuss the model features, training data, and loss in detail.

We have used disjoint data mode, which allows multiple input nets to be processed at the same time by creating a sparse diagonal block union of their adjacency matrices. Disjoint mode is visualized in Fig.~\ref{fig:model_structure} with two nets. This approach allows training and inference for multiple nets to run in parallel and improves runtime significantly.

\subsection{Model Features and Labels}


In order to extract the topographical features of the net, we use the Hanan grid of the net's pins (e.g., blue squares) as the input graph as seen in Fig.~\ref{fig:gnn-message-pass} and Fig.~\ref{fig:model_structure}. Empty nodes (e.g., purple circles) in the input graph can be chosen as the Steiner nodes (e.g., green circle). Features for every node are made up of their $x$ and $y$ coordinates and identification of node type, which can be a pin or an empty node.

An adjacency matrix is constructed representing each of the four orthogonal directions that connect sinks and empty nodes in the Hanan grid. 

The true output labels are binary node classifications which identify the correct, optimal Steiner nodes.

\subsection{Labeled Training Data}

GeoSteiner produces optimal results for RSMT problems but can take excessive run-time for large instances. We use GeoSteiner to label training data for up to degree 50 nets. 

In order to make sure the model learns from as many different RSMT examples as possible and the model scales for large degree nets, we generated a synthetic dataset of random nets. All coordinates are chosen randomly between 0 and 1,000,000. The dimensions of each problem instance are normalized to floating point numbers between 0 and 100. We found that, when normalized, the training samples produce more consistent patterns that the model can learn.

Table~\ref{table:training_dataset_stats} shows statistics for the randomly generated training data. We randomly generate 1,000 nets for each degree between 3 and 50 which makes training uniform across net degrees. We do not use nets of degrees larger than 50 in training. We use standard techniques for training using the sampled random data by selecting 80\% training and 10\% validation data. We collected the final accuracy of the training phase using the remaining 10\% as test data, but perform evaluation using more extensive, separate datasets in Section~\ref{sec:results}.

\begin{table}[ht]
\centering
\def\arraystretch{1.5}
\caption{Training Dataset Statistics}
\label{table:training_dataset_stats}
\centering
\setlength\tabcolsep{5pt}
\begin{tabular}{|c|c|c|c|c|c|c|c|}
\hline
\multirow{2}{*}{\shortstack{\# of nets\\($\times10^3$)}}& \multicolumn{6}{c|}{\# of nets by degree ($\times10^3$)} & \multirow{2}{*}{\shortstack{Degree of\\largest net}} \\
\cline{2-7}
     & 3-9 & 10-19 & 20-29 & 30-39 & 40-49 & 50 & \\
\hline
48  & 7   & 10    & 10    & 10    & 10    & 1         & 50 \\
\hline
\end{tabular}
\end{table}

\subsection{Training Evaluation}
\label{sec:training_evaluation}
We use Binary Focal Loss (BFL)~\cite{binaryfocalloss} which weighs the less occurring labels higher to train the model better. In our case, there are $\approx$6\% Steiner nodes and $\approx$94\% non-Steiner nodes. This made it so that when it was trained with unweighted binary cross-entropy, the model would always predict non-Steiner for all nodes. On the other hand, BFL can be computed as
\begin{equation}
\label{bfl}
    BFL(p) = -\alpha(1 - p)^\gamma \log(p)
\end{equation}
where $\alpha$ is the class balancing factor for class 1 (Steiner node); $\gamma$ is the focal factor and $p$ is the model predicted probability. We set the $\alpha$ value to 0.8, $\gamma$ value to 2, and used summation reduction.

We also implemented a custom confusion matrix to use as the model's accuracy metric. This confusion matrix ignores accurate predictions of non-Steiner nodes (true negatives). Since our labeled data had $\approx$94\% non-Steiner nodes, taking true negatives into account would generate a misleading accuracy. The custom confusion matrix uses the following formula
\begin{equation}
Accuracy = 
\begin{cases}
    \qquad\>\>\>1, & \text{if } TP + FP + FN = 0\\
    \frac{TP}{ (TP + FP + FN) }, & \text{otherwise}
\end{cases}
\label{eq:accuracy}
\end{equation}
where $TP$ is the correct prediction of Steiner nodes (true positives), $FP$ is the incorrect prediction of Steiner nodes (false positives), and $FN$ is the incorrect prediction of non-Steiner nodes (false negatives). Some low degree problems do not have any Steiner nodes since the original nodes align perfectly. We assume these cases are correct predictions if the model predicts all non-Steiner nodes. We used this metric for all accuracy reported in this paper.

\subsection{Hyperparameter Tuning}

We tuned model hyperparameters with Keras Tuner~\cite{kerastuner} using the range of values in Table~\ref{table:model_parameters} and the Hyperband tuner~\cite{hyperband} with Binary Focal Loss in Equation~\ref{bfl} on the validation set. We found the model with the highest accuracy for the validation data had \modelNumConvLayers{} GAT convolution layers. The first used ELU activation, \modelNumChannels{} channels and \modelNumAttentionHeads{} attention heads. The final layer used sigmoid activation and had a fixed number of channels and number of attention heads of 1 in order to produce a single probability value per node. 


\begin{table}[ht]
\centering
\def\arraystretch{1.5}
\caption{Model parameters}
\label{table:model_parameters}
\setlength\tabcolsep{6pt}
\begin{tabular}{|c|c|c|}
\hline
Parameter                                       & Range Explored & 
Best Value(s)\\
\hline
\multicolumn{1}{|l|}{Number of GAT layers}      & [2, 8] & 2 \\
\hline
\multicolumn{1}{|l|}{Number of channels}        & [2, 64] & 2$\rightarrow$1 \\
\hline
\multicolumn{1}{|l|}{Number of attention heads} & [1, 64] & 8$\rightarrow$1\\
\hline
\multicolumn{1}{|l|}{Attention dropout}         & [0.0, 0.25] & 0.225\\
\hline
\multicolumn{1}{|l|}{Layer dropout}             & [0.0, 0.25] & 0.0 \\
\hline
\end{tabular}
\end{table}

\subsection{Non-Steiner Refinement}
\label{sec:refinement}

\begin{figure}[tb] 
\begin{center}
\vspace{-0.3cm}
\includegraphics[width = 0.45\textwidth]{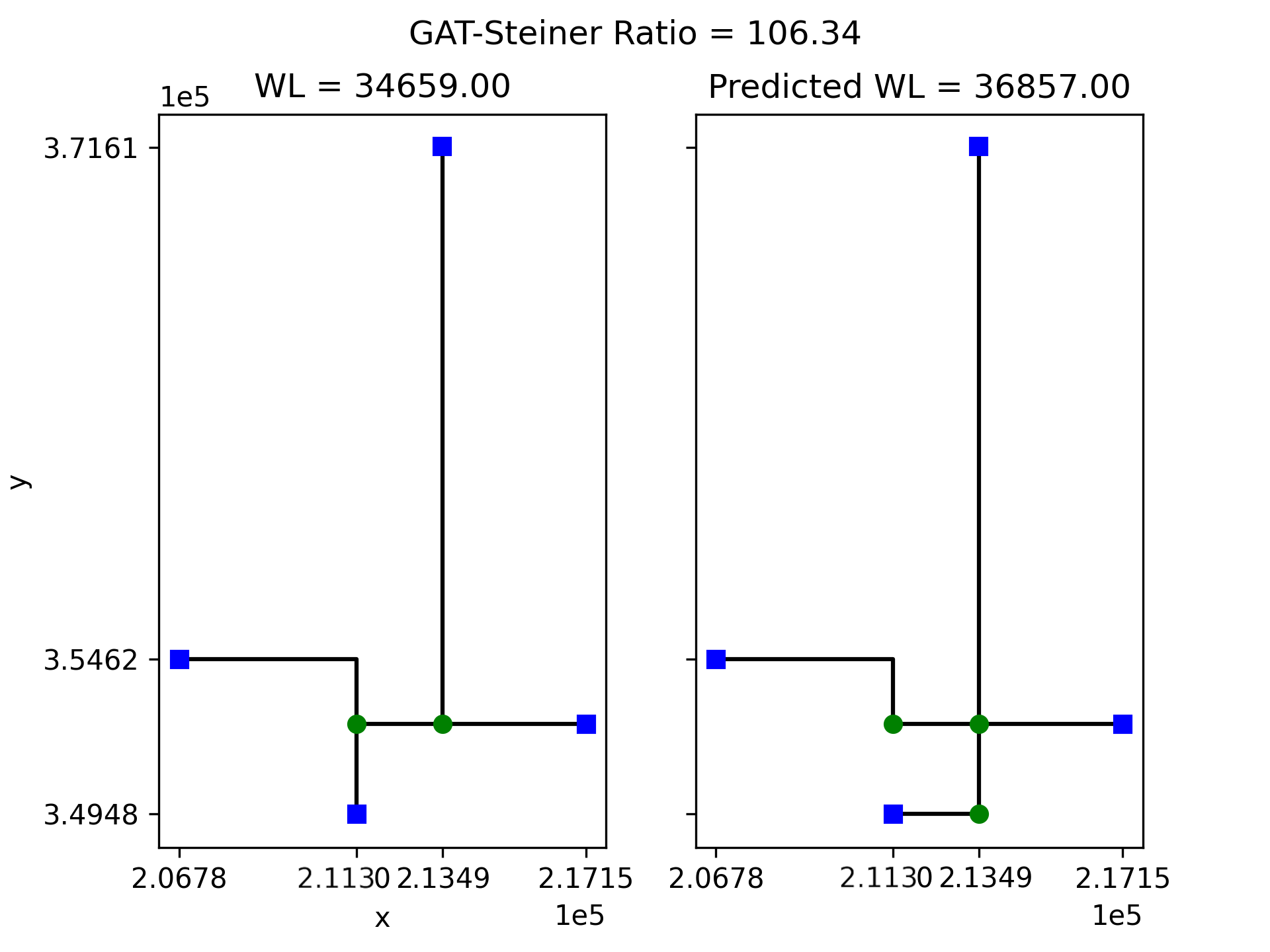}
 \caption {A net with a mispredicted degree-2 ``Steiner" node (which isn't by definition a Steiner node) that increases the total wire length by 6.34\%.
 \label{fig:extra_steiner}}
 \vspace{-0.3cm}
 \end{center}
\end{figure}

It is possible that we predict a node is a Steiner node when it is, in fact, not one. This would be obvious if, for example, the degree of the ``Steiner" node is only 2 as in Fig.~\ref{fig:extra_steiner}. We use quotes around the term Steiner in this case, because the nodes are predicted as Steiner nodes but are not technically Steiner nodes. For such cases, we developed a refinement strategy that is applied to all nets with mispredicted degree-2 ``Steiner" nodes:  

1. Consider all predicted Steiner nodes (of all node degrees) and remove the node with the lowest probability. Then, run the MST algorithm again. Keep doing step 1 until we no longer have degree-2 ``Steiner" nodes.

2. If the wire length has improved, return the last solution. Otherwise, recover the initial solution before step 1 and continue with step 3.

3. Consider only the degree-2 ``Steiner" nodes and remove the node with the lowest probability. Then, run the MST algorithm again. Keep doing step 3 until we no longer have degree-2 ``Steiner" nodes. Return the final solution.

In the worst case, this strategy will still have the same wire length as the initial solution. Step 3 cannot worsen the wire length since we are only removing degree-2 ``Steiner" nodes, which are unnecessary. 

Our non-Steiner refinement is different than other iterative Steiner prediction methods~\cite{delaware} because we only examine degree-2 nodes that are extremely rare whereas the other methods iteratively add one node at a time for every Steiner node, which requires multiple calls to the GNN model inference.

%% file: results.tex
\section{Results} 
\label{sec:results}

\subsection{Methodology}

We implemented our model using the Tensorflow~\cite{tensorflow} and Spektral~\cite{spektral}. We used the ADAM optimizer with a learning rate of 0.01 and early stopping with a patience of 5 epochs.

All serial data generation, testing, and evaluation are done on a server with two AMD EPYC 7542 32-core  2.9 GHz processors (128 threads total) and 512GiB DRAM. Programs were run on a single core of this machine, and we have not used a graphics processing unit (GPU) or tensor processing unit (TPU). Parallel training, testing, and evaluation are done with a NVIDIA GeForce RTX 4090 24GiB card. 

We used our serial-execution server for the other heuristics. We downloaded the source codes of FLUTE and SALT. FLUTE has accuracy and local refinement options. Accuracy is set to 3 by default, and can be 18 at maximum. Local refinement is suggested to be enabled if accuracy is larger than 4. We compiled FLUTE with two settings. FLUTE-3 is compiled with default options. FLUTE-18 is compiled with accuracy set to 18 and local refinement enabled. For both settings, we set the maximum degree to 3,000. SALT is compiled with the same FLUTE-3 settings and given an epsilon value of 10,000 so that it will optimize wire length instead of path length. We compiled GeoSteiner 5.3 with ILOG CPLEX Optimization Studio 12.6.3~\cite{cplex}, a linear programming solver library, for the best performance.

We used additional evaluation datasets from the ISPD 2019 routing contest benchmarks~\cite{ispd19} as well as  random nets that were not used in training. Our random evaluation nets include sizes greater than 50 whereas we only trained on nets up to degree 50. The random evaluation dataset has 900 random nets each for degrees 3 to 50, and 1,000 random nets for degrees 100, 200, 300, 400, 500, 1000, 2000, 3000. As can be seen in Table~\ref{table:evaluation_dataset_stats}, the random evaluation data has a larger spread over all net sizes compared to the ISPD19 dataset. ISPD19 only has a few nets with degree greater than 100.

\begin{table}[ht]
\centering
\def\arraystretch{1.5}
\caption{Evaluation Dataset Statistics}
\label{table:evaluation_dataset_stats}
\centering
\setlength\tabcolsep{1.5 pt}
\begin{tabular}{|c|c|c|c|c|c|c|c|c|c|}
\hline
\multirow{2}{*}{Dataset} & \multirow{2}{*}{\shortstack{\# of nets\\($\times10^3$)}}& \multicolumn{7}{c|}{Approx. \# of nets by degree ($\times10^3$)} & \multirow{2}{*}{\shortstack{Degree of\\largest net}} \\
\cline{3-9}
                                        &     & 3-9 & 10-19 & 20-29 & 30-39 & 40-49 & 50-99 & $\geq$100 & \\
\hline
\multicolumn{1}{|l|}{Random} & 440 & 63  & 90    & 90    & 90    & 90    & 9 & 8        & 3,000 \\
\hline
\multicolumn{1}{|l|}{ISPD19}            & 382 & 335 & 14    & 3     & 14    & 3     & 10 & 0.02       & 2,556 \\
\hline
\end{tabular}
\end{table}

\subsection{Accuracy Analysis}

The accuracy metric for the results is our custom confusion matrix in Eq.~\ref{eq:accuracy}. We generated solutions with GAT-Steiner, FLUTE, and SALT, and analyzed these predictions based on the optimal solutions generated by GeoSteiner. Note that our results regard label mispredictions with the same wire length as the correct predictions.

With our model, mispredictions are either a missing Steiner node or an extra Steiner node, with the majority of them being the latter due to the BFL weights in Section~\ref{sec:training_evaluation}. Fig.~\ref{fig:extra_steiner} illustrates an example of such suboptimal wire length net. Some problems might have multiple optimal solutions, or the predicted RSMT might predict a node as a Steiner node when it is not needed yet it does not affect the actual wire length of the net.

GAT-Steiner achieved \trainingAcc{} accuracy on the 10\% of training data reserved for testing, but we will do more thorough analysis in the rest of this section.

Table~\ref{tab:dataset_results} shows the average accuracy of Steiner point prediction for GAT-Steiner and the other heuristics on all nets. GAT-Steiner performs by far superior in the random evaluation over the other algorithms, since their heuristic approach does not perform well for large degree nets. It still performs slightly better on the ISPD19 dataset, which has a net degree distribution similar of that of a common design.

\begin{table}[tb]
\def\arraystretch{1.5}
\centering
\caption{Net Wire Length Results}
\setlength\tabcolsep{4pt}
\begin{tabular}{|l|c|c|c|c|}
\hline
\multicolumn{5}{|c|}{\textbf{Random Evaluation Dataset}} \\ \hline
                &  GAT-Steiner                 &  FLUTE-3                           & FLUTE-18                        &  SALT \\ \hline
Average accuracy    & \modelRandomAcc{}           & \fluteDefaultRandomAcc{}           & \fluteBestRandomAcc{}           & \saltRandomAcc{} \\ \hline
Suboptimal WL nets  & \modelRandomMispredictions{}& \fluteDefaultRandomMispredictions{}& \fluteBestRandomMispredictions{}& \saltRandomMispredictions{} \\ \hline
Average WL increase & \modelRandomAvgWLIncrease{} & \fluteDefaultRandomAvgWLIncrease{} & \fluteBestRandomAvgWLIncrease{} & \saltRandomAvgWLIncrease{} \\ \hline
Max WL increase & \modelRandomMaxWLIncrease{} & \fluteDefaultRandomMaxWLIncrease{} & \fluteBestRandomMaxWLIncrease{} & \saltRandomMaxWLIncrease{} \\ \hline

\multicolumn{5}{|c|}{\textbf{ISPD19 Dataset}} \\ \hline
                &  GAT-Steiner              &  FLUTE-3                         &  FLUTE-18                        &  SALT \\ \hline
Average accuracy    & \modelIspdAcc{}           & \fluteDefaultIspdAcc{}           & \fluteBestIspdAcc{}           & \saltIspdAcc{} \\ \hline
Suboptimal WL nets  & \modelIspdMispredictions{}& \fluteDefaultIspdMispredictions{}& \fluteBestIspdMispredictions{}& \saltIspdMispredictions{} \\ \hline
Average WL increase & \modelIspdAvgWLIncrease{} & \fluteDefaultIspdAvgWLIncrease{} & \fluteBestIspdAvgWLIncrease{} & \saltIspdAvgWLIncrease{} \\ \hline
Max WL increase & \modelIspdMaxWLIncrease{} & \fluteDefaultIspdMaxWLIncrease{} & \fluteBestIspdMaxWLIncrease{} & \saltIspdMaxWLIncrease{} \\ \hline
\end{tabular}
\label{tab:dataset_results}
\end{table}

\subsection{Suboptimality Analysis} 

We also did an analysis of the impact of mispredictions on solution quality. Table~\ref{tab:dataset_results} shows the rate of suboptimal wire length nets. Many papers only present the results including correctly predicted nets which can be misleading since many easy nets are easily predicted correctly (e.g., 3-pin nets or nets with no Steiner points) whereas harder nets can be far from optimal. Often, benchmarks are dominated by easy nets, which give a misleading picture of overall capability.

\begin{figure}[htb] 
\begin{center}
\vspace{-0.3cm}
\includegraphics[width = 0.45\textwidth]{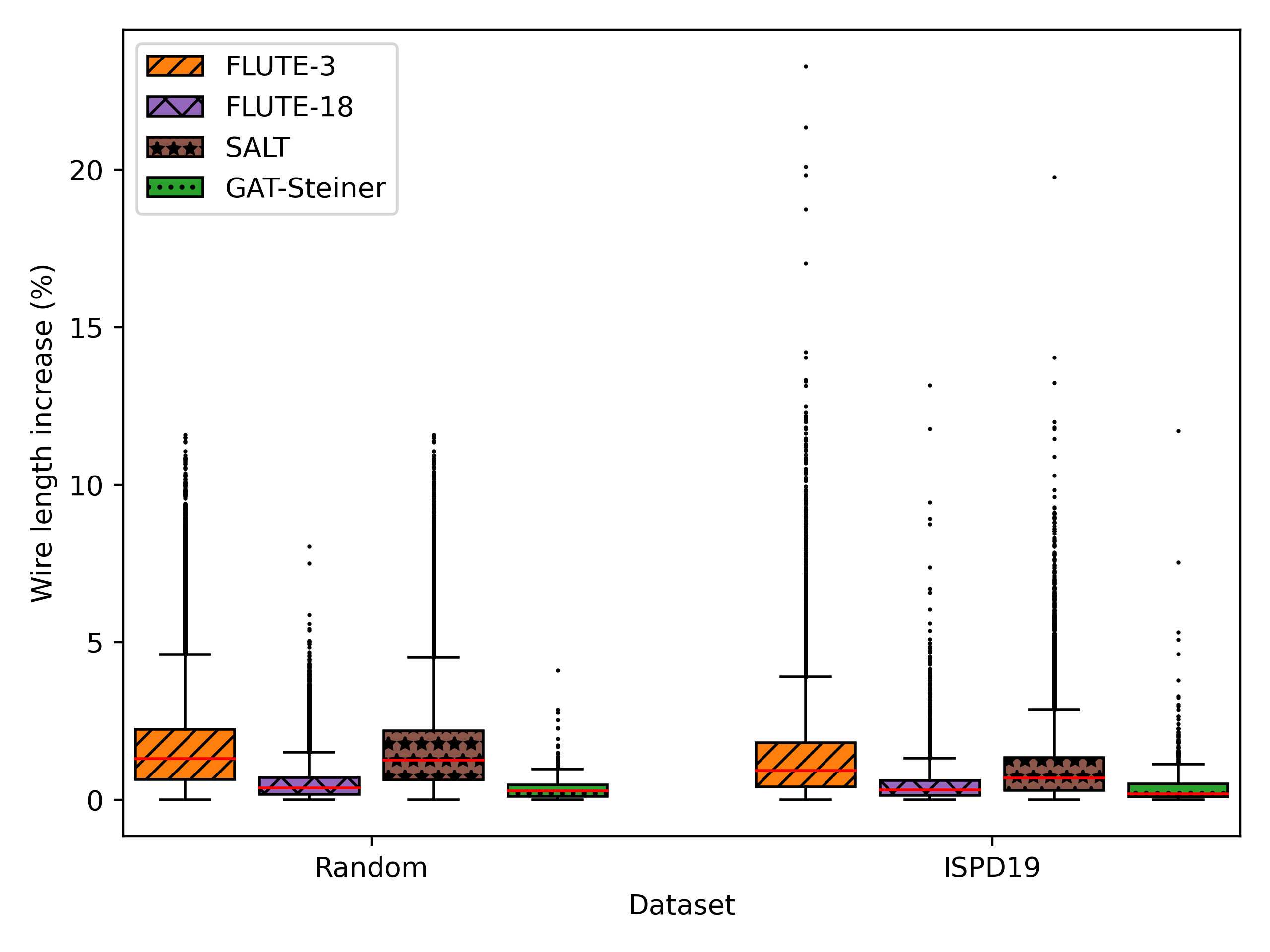}
 \caption {GAT-Steiner makes an average error of less than 1\% in wire length on suboptimal nets across all datasets and has the smallest maximum suboptimal wire length outlier. The red line shows the average wire length increase.
 \label{fig:wirelength}}
 \vspace{-0.3cm}
 \end{center}
\end{figure}

Fig.~\ref{fig:wirelength} shows the maximum, minimum and average wire length increase of suboptimal wire length instances from all heuristics. The average and maximum wire length increase is also in Table~\ref{tab:dataset_results}. The box and whisker plot defines an outlier as 1.5x the inter-quartile range (IQR) from the box. GAT-Steiner has the smallest wire-length increase in outliers and the fewest of them. Table~\ref{table:outliers} further shows that the number of outliers from GAT-Steiner is 1-2 orders of magnitude less than the other heuristic approaches.  

Since both FLUTE and SALT inherently work better on smaller instances, they have much worse accuracy on the random dataset which is distributed over a larger range of net sizes (Table~\ref{table:evaluation_dataset_stats}).

{
\begin{table}[htb]
\centering
\def\arraystretch{1.5}
\caption{Number of Outliers in Fig. \ref{fig:wirelength}}
\label{table:outliers}
\begin{tabular}{|l|c|c|c|c|}
\hline
       & FLUTE-3 & FLUTE-18 & SALT  & GAT-Steiner \\ \hline
Random & 9,741   & 6,098    & 9,747 & 23          \\ \hline
ISPD19 & 1,322   & 547      & 1,121 & 70          \\ \hline
\end{tabular}
\end{table}
}


GAT-Steiner produced degree-2 nodes predicted as ``Steiner" nodes for \modelRandomRanRefinement{} and \modelIspdRanRefinement{} of the nets in the random dataset and the ISPD19 dataset, respectively. On nets with these nodes, we ran our refinement algorithm from Section~\ref{sec:refinement}. \modelRandomAlreadyOptimalPercentage{} and \modelIspdAlreadyOptimalPercentage{} of these nodes were already the optimal solutions and therefore removing the unnecessary degree-2 nodes did not alter the solution. This is because the extraneous degree-2 nodes are usually on the optimal path and do not affect the actual wire length. Since this refinement step is only run for a small percentage of nets, its runtime overhead was negligible.


\begin{figure}[htb] 
\begin{center}
\vspace{-0.3cm}
\includegraphics[width = 0.45\textwidth]{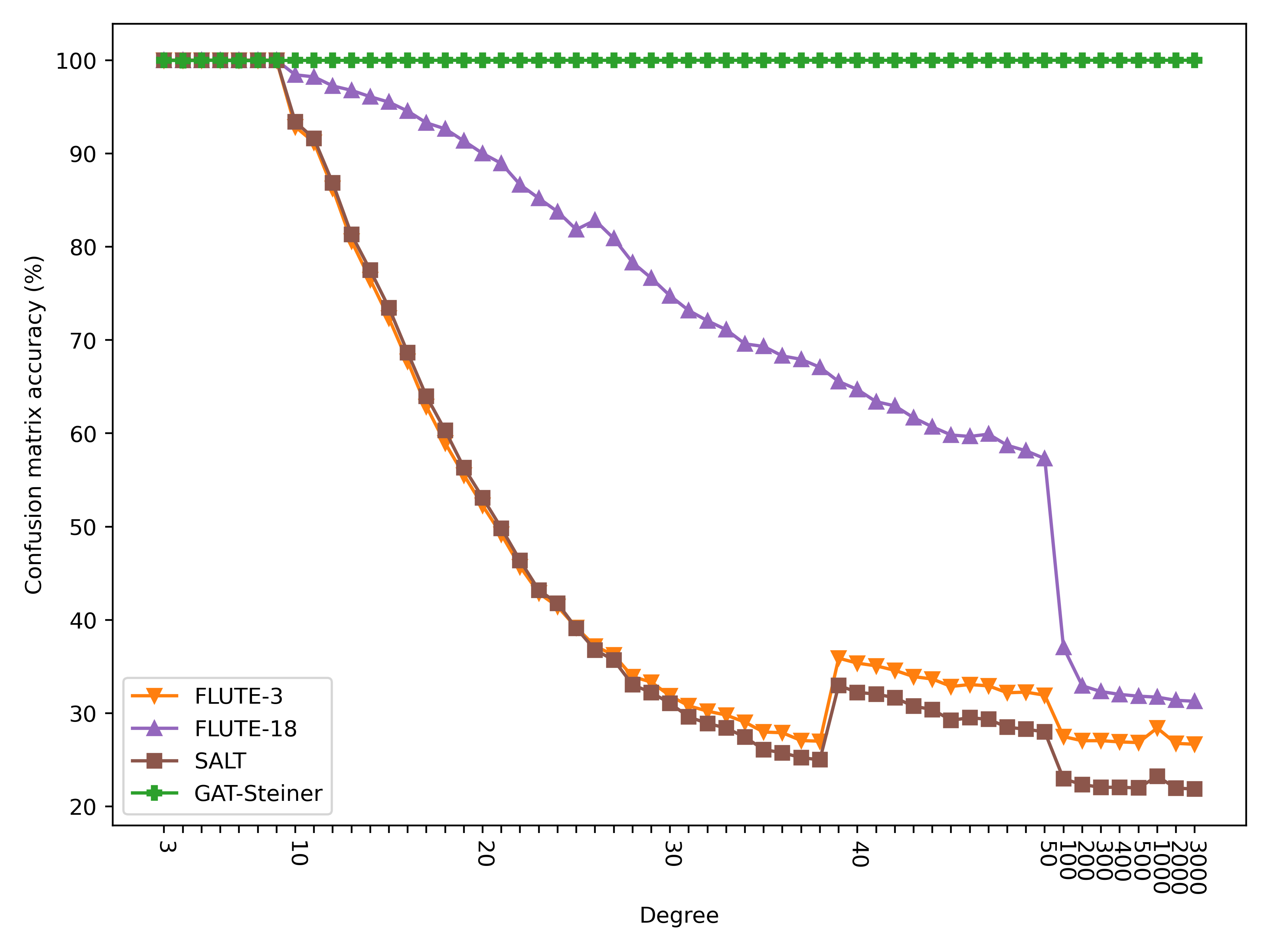}
 \caption {GAT-Steiner's accuracy remains above 99\% as problem size scales up while other heuristics' accuracy decline.
 \label{fig:degree_accuracy}}
 \vspace{-0.3cm}
 \end{center}
\end{figure}

\subsection{Parallelization Analysis}

GeoSteiner, FLUTE, and SALT are implemented in single-threaded C/C++ programs; therefore, they can only solve problems sequentially. Our model is implemented in Python, which is slower than C/C++, but it can solve problems in parallel using our GNN model on a GPU. Although our model might be slower than GeoSteiner when only a single RSMT is solved, it has significant speedups when RSMT instances are solved in parallel.

We used the random evaluation dataset to measure speedup since we have more uniform data across degrees. We used the largest batch size that fit into the GPU's memory. For nets having degree 3 to 50, 1,000 nets fit into the memory. For larger nets, we had to scale our batch size down. The batch size affects the speed proportionally; therefore, it can be set higher if GPU memory allows.

\begin{figure}[htb] 
\begin{center}
\vspace{-0.3cm}
\includegraphics[width = 0.45\textwidth]{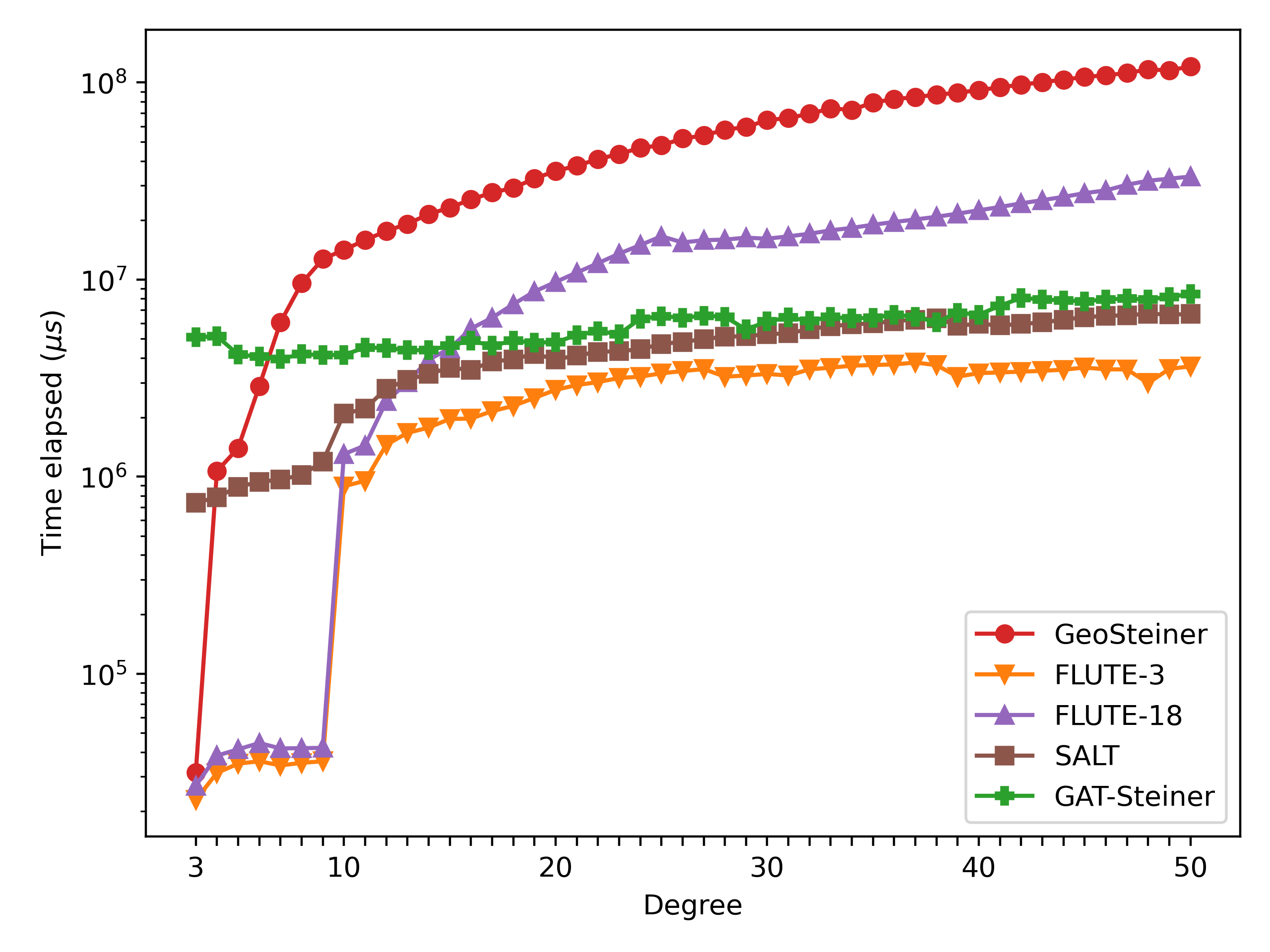}
 \caption {GAT-Steiner has the advantage of parallelism for larger instances compared to the other algorithms.
 \label{fig:time_1000}}
 \vspace{-0.3cm}
 \end{center}
\end{figure}

\begin{figure}[htb] 
\begin{center}
\vspace{-0.3cm}
\includegraphics[width = 0.45\textwidth]{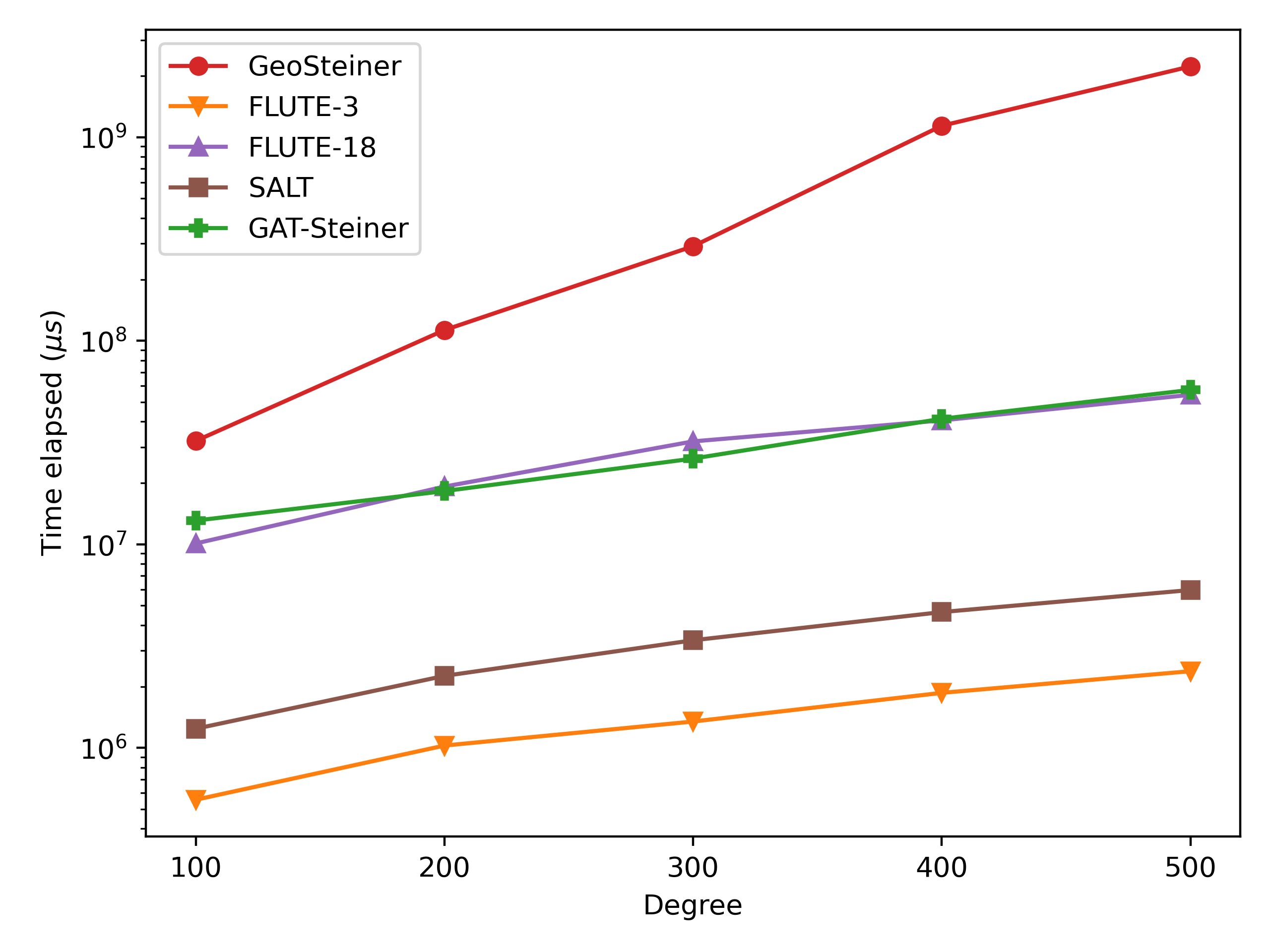}
 \caption {GAT-Steiner has a parallelism advantage for large instances even if parallelism needs to be scaled down due to GPU memory limits.
 \label{fig:time_50}}
 \vspace{-0.3cm}
 \end{center}
\end{figure}

Fig.~\ref{fig:time_1000} and Fig.~\ref{fig:time_50} show the total GPU execution time of all batches using the random evaluation dataset. GeoSteiner, FLUTE, and SALT are run sequentially. For the results shown in Fig.~\ref{fig:time_1000}, the average speedup of GAT-Steiner is \modelSpeedupLargeBatch{} over GeoSteiner. GAT-Steiner achieved an average speedup of \modelSpeedupSmallBatch{} for the subset of Fig.~\ref{fig:time_50}. For the ISPD19 dataset, GAT-Steiner was able to run inference with only 4 batches of size 100,000 total nets each.

\subsection{Hyperparameter Analysis}

Using the best model according to the Keras tuner result in Table~\ref{table:model_parameters} as a baseline, we explored the effects of each hyperparameter in our model separately by performing a sensitivity analysis of each with the confusion matrix accuracy (i.e., fixing every other parameter and then examining how the accuracy of our model is affected). 

From our analysis, we observed that beyond the optimal 2-layer configuration, the accuracy significantly drops. This is in line with our expectations, since we expect over-smoothing to be a factor with increasing number of layers. Luckily, however, deep networks are unnecessary for even very large nets as shown earlier.

The number of channels and the number of attention heads do not significantly affect the accuracy and result in changes of at most 0.2\%. For both parameters, the accuracy almost reaches our optimal model's with 2 channels and 2 attention heads and beyond this point the accuracy mostly stays within a $\pm$0.05\% range. This provides insight that GAT-Steiner reaches a sufficient number of learnable parameters with a relatively small number of channels and attention heads. This also confirms (in addition to our evaluation experiments) that we are not over-fitting the data.

Our model did not benefit from the use of feature dropout layers. We observed that with dropout rates of just 2.5\%, our model would lose about 5\% accuracy. We believe that this might be due to our model not having many input features (i.e., just $x$ location, $y$ location, and node type). Since every feature is critical to determining if a node is a Steiner node, feature dropout had a negative effect on our model's accuracy.  

On the other hand, attention dropout is a similar mechanism except that a neighbor's effect on a node can be dropped out completely. In particular, attention dropout prevents a node from relying on obtaining features through a particular path in the graph to other nodes. Instead, attention dropout requires that a node learn about nearby nodes and which features are shared through multiple paths simultaneously. Attention dropout does affect overall accuracy and enables the model to learn a more robust set of weights. However, we saw that at attention dropout rates greater than 30\%, the model starts to mispredict more often.

%% file: conclusion.tex
\section{Conclusion}
\label{sec:conclusion}
In this paper, we proposed GAT-Steiner, a graph attention network model to predict Steiner points for the Rectilinear Steiner Minimal Tree (RSMT) problems. GAT-Steiner can be used to predict RSMTs in bulk with very high accuracy and 1-2 orders of magnitude fewer wire-length outliers than heuristic approaches. Our model achieved \modelRandomAcc{} accuracy on the randomly generated test data and \modelIspdAcc{} accuracy on the ISPD 2019 benchmarks.